\pdfoutput=1

\documentclass[11pt]{article}

\usepackage[preprint]{acl}

\usepackage{times}
\usepackage{latexsym}

\usepackage[T1]{fontenc}

\usepackage[utf8]{inputenc}

\usepackage{microtype}

\usepackage{inconsolata}

\usepackage{graphicx}

 \usepackage{fontawesome5}

\usepackage{microtype}
\usepackage{hyperref}
\usepackage{url}
\usepackage{booktabs}
\usepackage{graphicx}
\usepackage{pgfplots}
\usepackage{pgfplotstable}
\pgfplotsset{compat=newest}
\usetikzlibrary{pgfplots.groupplots}

\usepackage{subcaption}
\usepackage{enumitem}
\newenvironment{itemize*}%
 {\leftmargini=20pt\begin{itemize}%
  \setlength{\itemsep}{3pt}%
  \setlength{\parskip}{0pt}%
  }%
 {\end{itemize}}
\newenvironment{enumerate*}%
 {\begin{enumerate}%
  \setlength{\itemsep}{0pt}%
  \setlength{\parskip}{0pt}}%
 {\end{enumerate}}

\usepackage{listings}
\usepackage{titlesec}
\usepackage{multirow}
\usepackage{makecell}
\usepackage{xcolor}
\usepackage{fontawesome5}
\usepackage[most, breakable, many]{tcolorbox}
\usepackage{xspace}
\usepackage{cleveref}

\newcommand{\sysname}{\textsc{TD-Eval}}
\newcommand{\convarena}{TOD Agent Arena}

\definecolor{lightred}{RGB}{255,163,163}
\definecolor{deepred}{RGB}{146,0,0}

\definecolor{midnightgreen}{rgb}{0.0, 0.29, 0.33}
\definecolor{deepgreen}{HTML}{0aa344}
\definecolor{deeppurple}{HTML}{7030a0}
\definecolor{deepblue}{HTML}{171d91}
\definecolor{brown}{HTML}{843c0c}
\definecolor{shadered}{HTML}{ffe5e5}
\definecolor{shadegreen}{HTML}{e5f7ed}

\title{\faSync \: \sysname: Revisiting Task-Oriented Dialogue Evaluation by Combining Turn-Level Precision with Dialogue-Level Comparisons}

\author{
Emre Can Acikgoz\thanks{\ \ indicates equal contribution.}, Carl Guo$^{*}$, Suvodip Dey$^{*}$, Akul Datta, Takyoung Kim,\\
\textbf{Gokhan Tur, Dilek Hakkani-Tür}\\
University of Illinois Urbana-Champaign\\
\texttt{\{acikgoz2, carlguo2, sdey, gokhan, dilek\}@illinois.edu}\\
}

\begin{document}
\maketitle

\begin{abstract}
Task-oriented dialogue (TOD) systems are experiencing a revolution driven by Large Language Models (LLMs), yet the evaluation methodologies for these systems remain insufficient for their growing sophistication.
While traditional automatic metrics effectively assessed earlier modular systems, they focus solely on the dialogue level and cannot detect critical intermediate errors that can arise during user-agent interactions.
In this paper, we introduce \sysname{} (\textbf{T}urn and \textbf{D}ialogue-level \textbf{Eval}uation), a two-step evaluation framework that unifies fine-grained turn-level analysis with holistic dialogue-level comparisons. 
At turn level, we evaluate each response along three TOD-specific dimensions: \textit{conversation cohesion}, \textit{backend knowledge consistency}, and \textit{policy compliance}. 
Meanwhile, we design \textit{\convarena{}} that uses pairwise comparisons to provide a measure of dialogue-level quality.
Through experiments on MultiWOZ 2.4 and $\tau$-Bench, we demonstrate that \sysname{} effectively identifies the conversational errors that conventional metrics miss.
Furthermore, \sysname{} exhibits better alignment with human judgments than traditional and LLM-based metrics.
These findings demonstrate that \sysname{} introduces a new paradigm for TOD system evaluation, efficiently assessing both turn and system levels with a plug-and-play framework for future research\footnote{Project Page: \url{https://emrecanacikgoz.github.io/TD-Eval/}}.

\end{abstract}

\section{Introduction}
Task-oriented dialogue (TOD) systems are conversational agents that help users complete specific tasks such as booking hotels, ordering food, or scheduling appointments. 
Advances in Large Language Models (LLMs) have significantly enhanced the capabilities and flexibility of modern TOD systems~\citep{hudecek2023arellmstod, acikgoz2025desideratum}. 
However, evaluating their true conversational capabilities remains a challenge~\citep{nekvinda2021shadesbleu}. Although human evaluation serves as the gold standard for evaluating dialogue systems, conducting tests with real users is time-consuming, costly, and difficult to scale across multiple systems and iterations. This limitation creates a significant gap in measuring and ensuring accountability in TOD systems research.

\begin{figure}[t!]
\includegraphics[width=\linewidth]{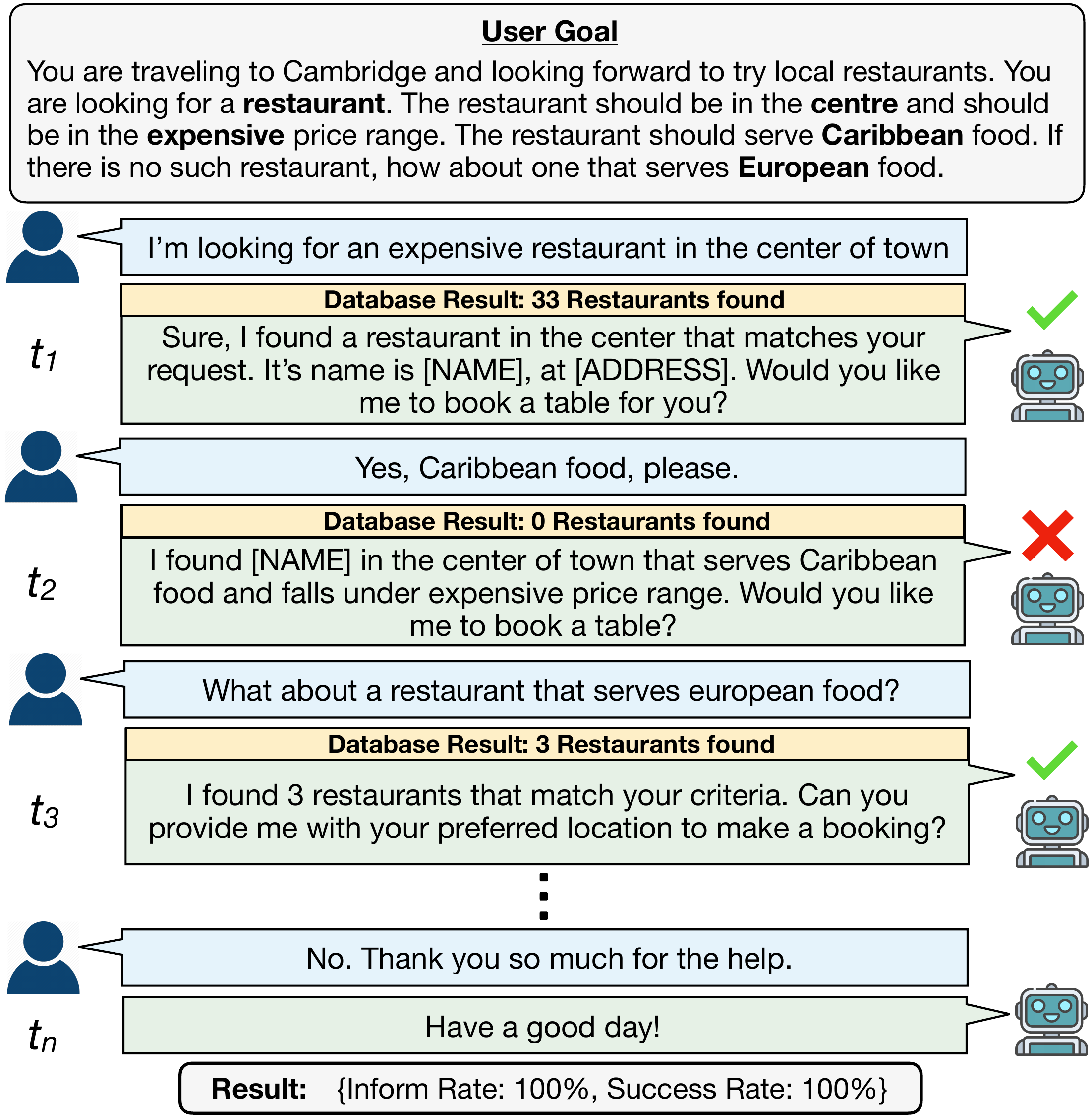}
\caption{Sample conversation between a user and TOD agent with a MultiWOZ goal. Commonly used evaluation metrics in TOD, Inform and Success, fail to detect the turn-level error and assign a perfect score.
}
\vspace{-7mm}
\label{fig:dialog-example}
\end{figure}
 
To automate TOD evaluation, previous work has converged to various offline metrics. For example, metrics like \textit{Inform} and \textit{Success} rates are utilized to estimate task completion in MultiWOZ~\citep{budzianowski2018multiwoz}. Traditionally, the overall performance of TOD systems is evaluated at the dialogue-level using task completion metrics, ignoring turn-level performance. While a few metrics assess turn-level quality. For example, Joint Goal Accuracy (JGA)~\citep{budzianowski2018multiwoz} for dialogue understanding and BLEU~\citep{papineni2022bleu} for response generation. However, these metrics rely heavily on ground-truth annotations, making them impractical for evaluating arbitrary conversations without labeled data.

In typical TOD systems, each user turn is processed by converting the dialogue context into a query to a backend database. The system then decides on the next response by combining these database results with the user’s request. However, as discussed, commonly used TOD metrics only capture the summative outcome of a dialogue and may not penalize intermediate system misdirections. For example, in \Cref{fig:dialog-example}, the system hallucinates in the second turn ($t_2$) and misinforms the user, suggesting that there is a restaurant that matches the user's request despite the empty response from the database. However, the mistake is overwritten in subsequent turns, resulting in 100\% Inform and Success (indicating perfect task completion), despite the fact that an intermediate step is completely wrong.
On the other hand, task completion relies on delexicalizing responses by replacing entity attributes with placeholders (e.g., [NAME], [ADDRESS])~\citep{budzianowski2018multiwoz}. This obscures critical errors (e.g., incorrect hotel names) while preserving the slot structure. Moreover, classic turn-level evaluations often depend on manual annotation, which is costly and time consuming. As underscored by recent studies~\citep{li2024fnctod, xu2024autotod}, more realistic and comprehensive evaluation approaches are required in the LLM era to capture both intermediate correctness and overall response quality, exposing how current automatic metrics often miss critical errors in TOD.

In this work, we propose \sysname{} (\textbf{T}urn and \textbf{D}ialogue-level \textbf{Eval}uation), an easy-to-use evaluation framework that combines turn-level performance with overall dialogue-level response comparisons as illustrated in \Cref{fig:main}. By integrating both levels, \sysname{} enables local error analysis for troubleshooting and global model-to-model comparisons for reliable performance benchmarking, while providing a more reliable and human aligned evaluation of conversational quality compared to other metrics. \sysname{} introduces three turn-level metrics: \emph{conversation cohesion}, \emph{backend knowledge consistency}, and \emph{policy compliance}, evaluated using an LLM judge \textit{together with its justifications}. These metrics help identify subtle errors often missed by traditional automatic evaluation methods. The LLM judge model is flexible and can be easily configured using any open-source or proprietary model, with GPT-4o~\citep{hurst2024gpt4o} as the default. The proposed framework complements the per-turn judgments with a dialogue-level comparison method: a \emph{\convarena{}}, where entire dialogues from competing agents are systematically ranked via an Elo-based pairwise evaluation. Unlike general-purpose chatbot arenas~\citep{zheng2023llmasjudge} that focus on open-domain or chit-chat dialogues, \convarena{} emphasizes task-oriented interactions requiring domain-specific database integration, measuring success by the agent’s ability to fulfill service-oriented goals rather than linguistic fluency.

The main contributions of our work are summarized as follows:
\begin{itemize}[topsep=2pt, partopsep=-5pt, leftmargin=8pt, itemsep=-4.5pt]
    \item We propose \sysname{}, a two-level framework that combines three turn-level metrics with pairwise comparisons at the dialogue level.
    \item Our framework uses LLM-based judging, capturing domain-specific errors overlooked by standard automatic metrics (e.g., Inform, Success).
    \item We demonstrate the effectiveness of \sysname{} on MultiWOZ 2.4 and $\tau$-Bench, validating with human evaluations.
    \item We perform a comprehensive evaluation of state-of-the-art LLMs on \sysname{} and provide insights into their strengths and limitations. We publicly release our code, system-response data, and human evaluations, along with a Huggingface Leaderboard for TOD Agents. 
\end{itemize}

\begin{figure*}[!t]
\includegraphics[width=\linewidth]{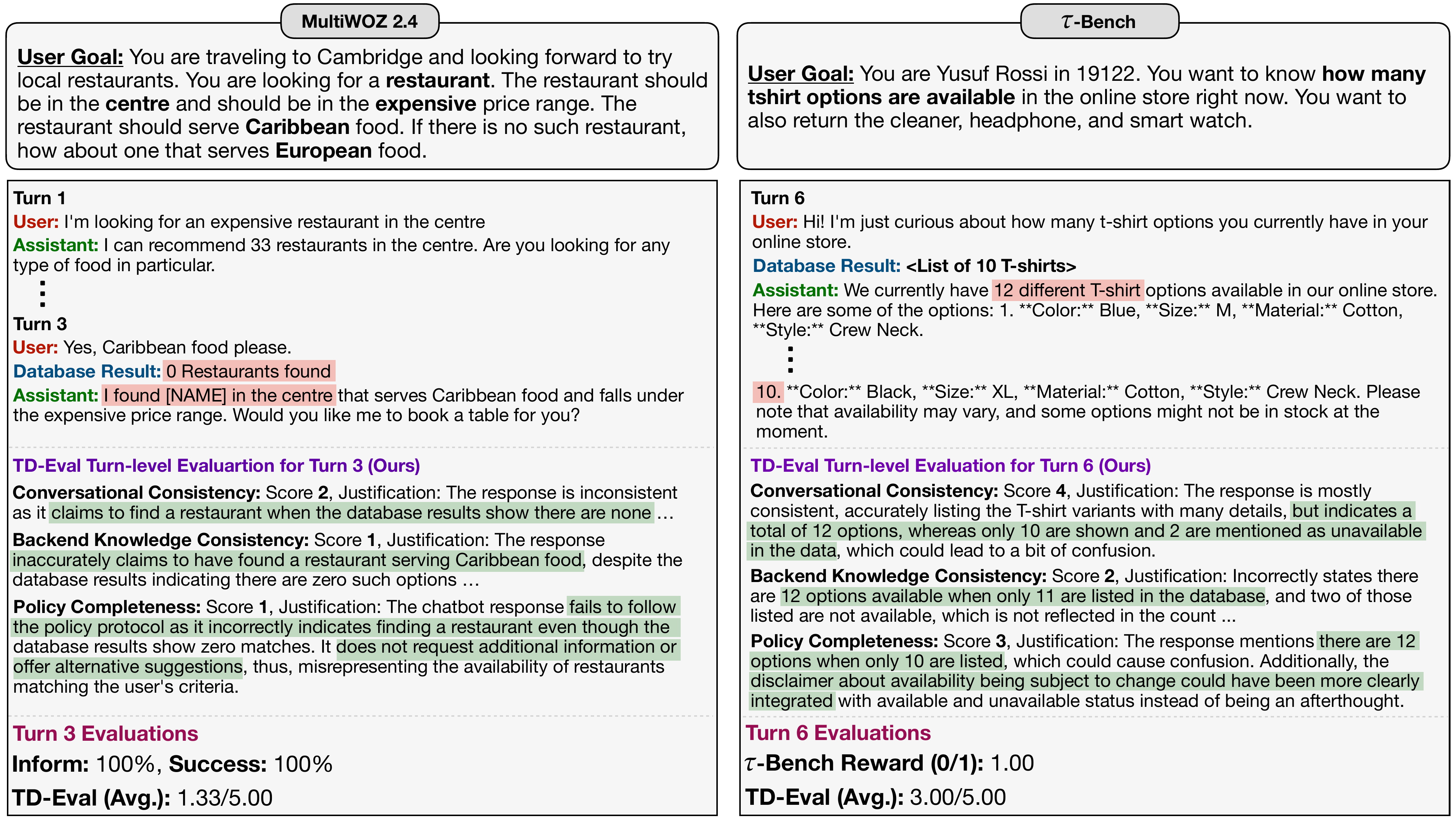}
\caption{
 \textbf{Comparison of \sysname{} with Automatic Evaluations.} \textbf{Left:}  In MultiWOZ 2.4, a non-existent restaurant is suggested but received a 100\% score from Inform and Success, while \sysname{}’s turn-level analysis assigns a lower score. \textbf{Right:} On $\tau$-Bench, the agent claims 12 T-shirts (only 10 exist), yet string matching treats it as correct; \sysname{} flags this mismatch. We highlight errors in \textbf{\textcolor{lightred}{red}} and correct elements in \textbf{\textcolor{deepgreen}{green}}.
}
\label{fig:td-eval-example}
\vspace{-5mm}
\end{figure*}

\section{Preliminaries}
\label{sec:preliminaries}
While various aspects can contribute to the \textit{satisfactory} evaluation of dialogue systems, we focus on task completion scenarios, reviewing \textbf{automatic metrics} used in TOD and analyzing their limitations in the context of modern dialogue systems.

\subsection{Automatic Metrics in TOD Evaluation}
Traditionally, the responses generated by TOD systems are evaluated primarily for task completion. MultiWOZ~\citep{budzianowski2018multiwoz, ye2022multiwoz24} contains two automatic task completion metrics, \textbf{Inform} and \textbf{Success}. In $\tau$-Bench~\citep{yao2025taubench}, authors defined a binary \textbf{Reward} metric that checks the action and output correctness.

\vspace{1mm}

\noindent\textbf{Inform and Success.} 
Inform measures the system's ability to provide correct entities from database search results in response to user requests. In addition to entity correctness, Success checks whether the system returns all the requested attributes. For each dialogue, Inform and Success are reported as binary scores (1 or 0), where 1 denotes success and 0 denotes failure. For example, if a user asks to \texttt{book a table at a suggested restaurant} and the system fails to confirm the reservation, the Success for that dialogue is 0.

\vspace{1mm}

\noindent\textbf{$\tau$-Bench Reward.}  $\tau$-Bench~\citep{yao2025taubench} defined a binary reward metric that captures both action correctness and output completeness. Specifically, let $r_{\text{action}} \in \{0,1\}$ be an indicator that the final state of the database matches the unique outcome of ground truth, and let $r_{\text{output}} \in \{0,1\}$ be an indicator that the system’s final response to the user contains all the required information. Then, the overall reward is computed as $r = r_{\text{action}} \times r_{\text{output}} \in \{0,1\}$. A successful dialogue must both perform the correct database updates (e.g., returning the correct items) and provide all necessary details (e.g., item prices) to the user.

\subsection{Limitations of Traditional Metrics}

\noindent\textbf{Lack of Turn-Level Granularity.} Inform, Success, and $\tau$-Bench rewards are calculated once the dialogue concludes and can miss specifics about how the system performs at each turn. As a result, errors early in the conversation are not penalized if the system eventually corrects itself. For instance, as illustrated in \Cref{fig:dialog-example} (left), a system might initially hallucinate a restaurant name but later rectify its mistake, resulting in a misleadingly high score. 

\vspace{1mm}

\noindent\textbf{Binary Nature and Oversimplification.} Each of these metrics treat correctness as an all-or-nothing concept. Any deviation from the exact requested entity is considered equally wrong, whether it is entirely incorrect (e.g., a different type of venue in a different location) or partially incorrect (e.g., correct area but wrong cuisine). This approach ignores the difference between small and large errors, which can pose challenges when evaluating LLM-based systems that may produce partially correct responses. Furthermore, LLMs exhibit a variety of failure modes that traditional systems do not. They may hallucinate details, misinterpret subtle intentions, or produce inconsistent responses across turns. Because these metrics focus only on the final outcome, they fail to capture these unique conversational flaws.

\begin{figure*}[!t]
\includegraphics[width=\linewidth]{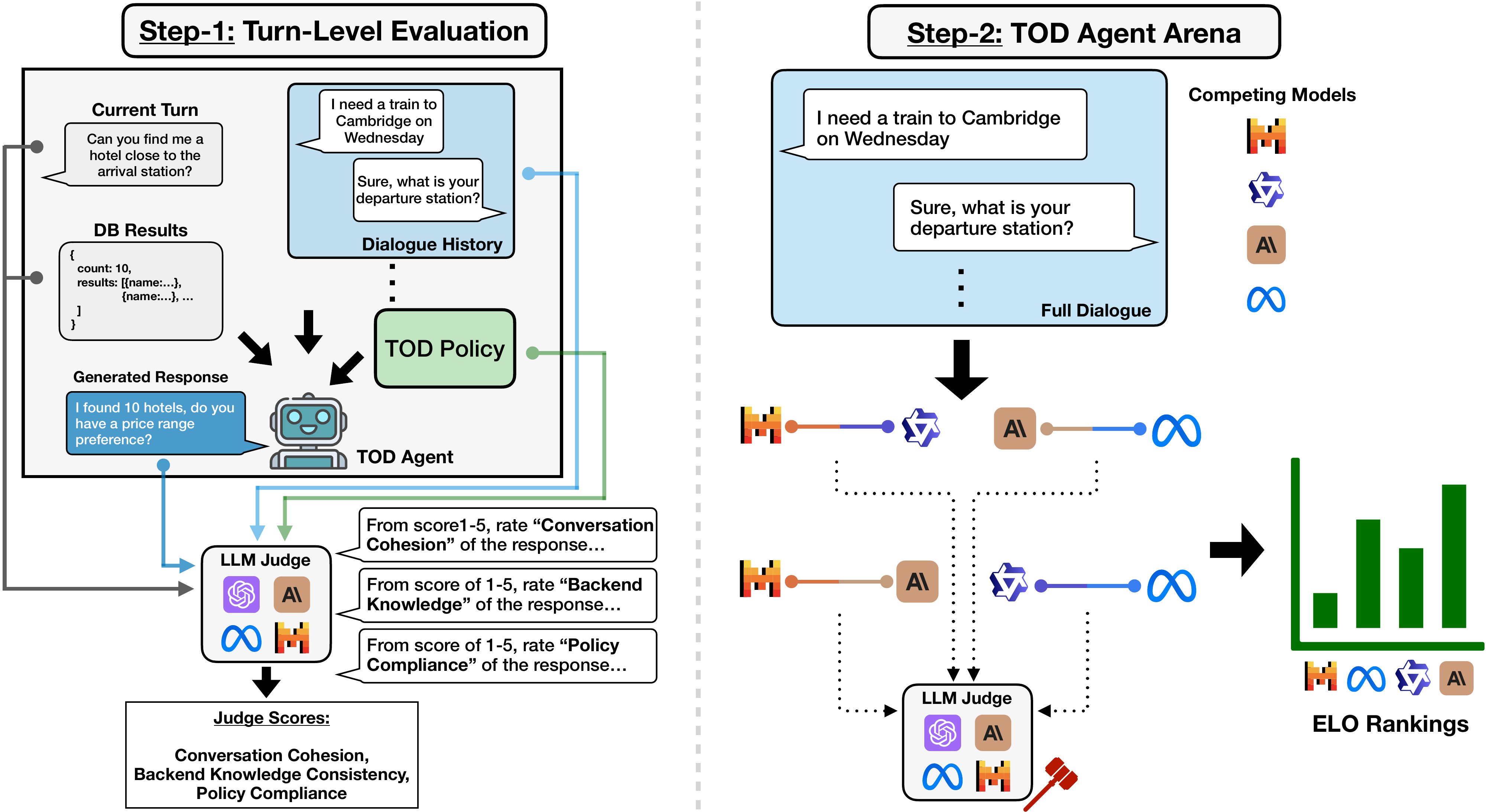}
\caption{\textbf{Overview of the \sysname{} Framework.} The left part illustrates the pipeline for turn-level evaluations, where a system is assessed across three criterion by an LLM-based judge. The right part depicts the second step, in which models compete within \convarena{} and ranked based on win and loss rates.}
\vspace{-5mm}
\label{fig:main}
\end{figure*}

\noindent\textbf{String-matching Vulnerabilities.} $\tau$-Bench's reward-based evaluations rely mainly on substring matching. These methods simply check for the presence of target substrings in model-generated responses, which can lead to misleading evaluations. For example, a model answering a numerical question incorrectly may still contain the correct number elsewhere in its response, either by coincidence or in an unrelated context, yet it still may be scored as correct like in \Cref{fig:td-eval-example} (right). 

\vspace{1mm}

These limitations collectively mean that automated metrics can provide incomplete or inflated estimates of true performance of TOD systems.

\section{\sysname{}}
\label{sec:td-eval}
\vspace{-2mm}

To comprehensively evaluate LLM performance on TOD tasks, it is critical to assess both turn-level and dialogue-level performances~\citep{siro2022usersatisfaction}. We propose a two-step evaluation protocol, \sysname{}, as shown in Figure \ref{fig:main}. 
The \textbf{first step} introduces a turn-level metric to evaluate TOD responses across three key dimensions: \textit{conversation cohesion}, \textit{backend knowledge consistency}, and \textit{policy compliance}. We chose these dimensions to specifically target the TOD domain. The \textbf{second step} extends this evaluation from turn to dialogue-level with a pairwise ranking method. For each dialogue pair, the LLM indicates a preference for the stronger conversation following prompted policies. \textit{This dual-step protocol offers a holistic view of TOD performance, capturing both localized (per turn) and global (whole-dialogue) qualities through a scalable, automated LLM-as-judge approach.} 

\vspace{-1mm}
\subsection{Turn-Level Evaluation}
\label{sec:turn-level}
As illustrated in the left-hand side of \Cref{fig:main}, turn-level evaluation focuses on analyzing the response of each system on each turn within the context of the dialogue, ensuring that intermediate errors are identified and penalized. The evaluation process involves feeding the user query, conversation history, database result, and system response into our framework, which uses a user-selected LLM as the primary evaluator. Each response is rated on a 5-point scale across the three defined dimensions, using prompts specifically designed to elicit accurate judgments.

\noindent\textbf{Conversation Cohesion.} This dimension evaluates how well the agent’s response aligns with the preceding dialogue context, ensuring the system remains relevant to both the user query and the dialogue history while maintaining a coherent flow. A key motivation for measuring conversation cohesion is to identify whether the agent remains on-topic and avoids illogical transitions, which are critical for a smooth user experience. To measure this, we prompt the LLM judge with the current conversation context (including the user query, dialogue history, database results, and the system’s response) and ask it to assign a score from 1 (Very Bad) to 5 (Very Good), along with its justification (See \Cref{fig:conversational-consistency}). 
\vspace{1mm}

\noindent\textbf{Backend Knowledge Consistency.} This dimension assesses how accurately the agent integrates external information (e.g., database results) into its responses, reflecting the system’s ability to retrieve and incorporate factual details. Maintaining correct backend knowledge is crucial to building user trust and ensuring that the agent can handle task-oriented requests (e.g., providing real-time data or service details) without misinformation. We feed the same conversation context and database results into the LLM judge with instructions (See \Cref{fig:backend-knowledge}).

\vspace{1mm}

\noindent\textbf{Policy Compliance.} A core requirement in TOD is following predefined, domain-specific policy protocols (e.g., when to request a user’s booking details or how many suggestions to offer). To evaluate this, we check whether an agent’s responses align with a policy prompt (\Cref{fig:policy-compliance}), covering the domain’s possible slots (e.g., time, price range), a set of policy rules (e.g., “wait for key information before making a reservation”), and a scoring guide. At each turn, the LLM judge references these instructions—along with the dialogue context and database outputs—to assign a discrete score. We create domain-specific prompts (e.g., for \textit{restaurant}, \textit{train}, \textit{hotel}, \textit{attraction}, \textit{taxi}), each reflecting unique policy objectives.
This structured approach ensures evaluations reveal how well the system follows the exact flows required in real-world services, addressing common failures of LLMs to adhere to detailed policies.

\subsection{Dialogue-Level Evaluation}
\label{sec:dialogue-level}

\noindent\textbf{Motivation for \convarena{}.} While turn-level metrics are useful for spotting localized issues, they can be too granular for capturing the overall user experience in extended conversations, especially in service-oriented conversations that require database (DB) interaction. For instance, if two models achieve close per-turn Likert scores, it can be difficult to discern which one ultimately yields a more coherent, user-friendly dialogue end-to-end. Moreover, turn-level assessments may not reflect human preferences in extended interactions. A slight advantage (e.g., an average score of 4.2 vs. 4.0) does not necessarily show how well the agent meets user goals or recovers from mistakes over multiple turns.

\vspace{1mm}

\noindent\textbf{Differences from Existing Arenas.} Unlike general-purpose chatbot arenas~\citep{zheng2023llmasjudge}, which often focus on open-domain or chit-chat dialogues, our \textbf{\convarena{}} specifically targets task-oriented scenarios that require integration with domain-specific DBs. This emphasis on DB-driven tasks introduces unique challenges: the system must retrieve, update, and reason about information stored in external data sources while maintaining coherent and contextually relevant multi-turn conversations. By centering our comparisons on these criteria, our arena offers a distinct perspective on conversational quality, where success is measured not only by linguistic fluency but also by the agent’s ability to fulfill service-oriented goals.

\vspace{1mm}

\noindent\textbf{Evaluation of Pairwise Comparisons with Elo.} We employ a pairwise ranking methodology inspired by MT-Bench~\citep{zheng2023llmasjudge}, which utilizes an Elo rating system. The Elo system provides a setup for evaluating relative performance through pairwise comparisons, as shown in the right side of \Cref{fig:main}. In our adaptation, all models begin with an initial rating of 1000, and their scores are dynamically adjusted based on LLM-based judgments of their responses. The process works as follows; first, two competing models generate responses for the same dialogue context as ``Conversation A'' and ``Conversation B''. Following the provided judge prompt (\Cref{fig:elo-prompt}), our LLM judge evaluates these conversations to determine between three options only: (i) ``Conversation A'' if Conversation A was better, (ii) ``Conversation B'' if Conversation B was better, or (iii) ``Equal'' if they were roughly equivalent. Based on this comparison, the models' Elo ratings are updated, with the magnitude of adjustment reflecting the outcome's predictability—unexpected victories (such as a lower-rated model outperforming a higher-rated one) result in larger rating changes. Thus, the dialogue-level arena adds a broader perspective that complements turn-level scores, ensuring that multi-turn dynamics, user goals, and overall satisfaction are adequately captured in our evaluation.

\section{Human Evaluation}
\label{sec:human_eval}

\subsection{Human Evaluation Process}
We conducted a comprehensive two-step human evaluation to validate \sysname{} as an effective metric. The study involved 10 annotators with academic backgrounds in Computer Science at the graduate level. All annotators were proficient in English; several have prior experience in NLP research, making them well-suited for nuanced evaluation of TOD conversations. 
The study was conducted using fully synthetic conversations generated from MultiWOZ and $\tau$-Bench goals with a user simulator. The data generation process followed the experimental setup described in \Cref{sec:exp_set}.
Annotators were presented with the dialogue history and the corresponding database results, and were asked to provide both turn-level and dialogue-level ratings for the three sub-metrics (conversation cohesion, backend knowledge, and policy compliance) on a 5-point Likert scale, ranging from``Very Bad'' to ``Very Good''. Detailed evaluation guidelines is provided in \Cref{fig:eval-instructions}. Refer to Appendix~\ref{app:human_eval_setup} for further details.

In the first step of human evaluation, all 10 participants annotated the same 5 conversations (3 from MultiWOZ and 2 from $\tau$-Bench). We observed a strong skew in the collected ratings, with nearly 90\% of scores falling in the 4–5 range (see \Cref{fig:score-distribution}). To account for this prevalence bias, we used Gwet's AC1~\citep{gwet} and Randolph's $\kappa$~\citep{randolph-kappa} to calculate agreement, both of which offer more stable agreement measures than traditional $\kappa$ statistics in such imbalanced settings~\citep{convosense}. As shown in \Cref{tab: iaa}, the inter-annotator agreement scores are between 0.4 and 0.7, indicating a substantial level of agreement among annotators. For the second step, we followed up with 9 of the original annotators. Each annotator was assigned 10 distinct conversations (6 from MultiWOZ and 4 from $\tau$-Bench), totaling of 90 annotated dialogues covering 490 turns.

\begin{table}[t]
\centering
\begin{small}
\begin{tabular}{l r r r} 
\toprule
\textbf{Metric}  & \textbf{Gwet AC1}  & \textbf{Randolph's $\kappa$} \\ \midrule
Conversation Cohesion & 0.49 & 0.42 \\
Backend Knowledge & 0.66 & 0.63 \\ 
Policy Compliance & 0.64 & 0.58 \\ \midrule
Overall & 0.56 & 0.54\\
\bottomrule
\end{tabular}%
\caption{Inter-Annotator Agreement scores from the first step of human evaluation. ``Overall'' combines scores from all three sub-metrics.}
\vspace{-0.2in}
\label{tab: iaa}
\end{small}
\end{table}

\subsection{Baseline Metrics}

\subsubsection{Traditional Metrics}
For MultiWOZ, we use the traditional Success rate~\citep{budzianowski2018multiwoz}, which is a stricter measure that subsumes Inform. For $\tau$-Bench, we use $\tau$-Bench reward~\citep{yao2025taubench}.


\subsubsection{LLM-as-judge}
We use the state-of-the-art LMUnit~\citep{lmunit} metric as our LLM-as-judge baseline. 
LMUnit evaluates a dialogue response against natural language unit tests using a unified scoring model, outputting a score between 1 and 5. To ensure a fair comparison with \sysname{}, we designed natural language unit test cases that capture our three sub-metrics (conversation consistency, backend knowledge correctness, and policy adherence). LMUnit scores are obtained via an API call, which takes the dialogue history, system response, and corresponding unit tests as inputs. We hand-crafted unit tests for both the dialogue level and the three turn level sub-metrics, shown in \Cref{fig:lmunit-prompts}. The turn-level sub-metrics are calculated as the mean score of the corresponding unit tests.
We refer to this extension of LMUnit as LMUnit\textsubscript{TD}.
Further details on the LMUnit test cases and API calls are provided in \Cref{app:human_eval_lmunit}. 

\begin{table}[!t]
\centering
\resizebox{\linewidth}{!}{%

\begin{tabular}{l l c c} 
\toprule
\textbf{Granularity} & \textbf{Metric}  & \textbf{Gwet AC1}  & \textbf{Randolph's $\kappa$} \\ \midrule
\multirow{2}{*}{Turn-level}
& LMUnit\textsubscript{TD} & 0.43 & 0.41  \\
& TD-EVAL & 0.56 & 0.50 \\ 
\midrule
\multirow{3}{*}{Dialogue-level}
& Traditional & 0.41 & 0.34 \\
& LMUnit\textsubscript{TD} & 0.44 & 0.40 \\
& TD-EVAL & 0.57 & 0.52 \\
\bottomrule
\end{tabular}%
 }
\caption{Comparison of different evaluation methods with human scores, using combined "Overall" score from all three metrics.}
\vspace{-0.2in}
\label{tab:agreements_final}
\end{table}

\begin{table*}[!t]
\centering
\resizebox{\textwidth}{!}{%
\begin{tabular}{lcccccccccc}
\toprule
\textbf{Model}                  & \multicolumn{3}{c}{\textbf{Conversational Consistency}} & \multicolumn{3}{c}{\textbf{Backend Knowledge}} & \multicolumn{3}{c}{\textbf{Policy Compliance}} & \textbf{Overall}  \\ 
\cmidrule(lr){2-4} \cmidrule(lr){5-7} \cmidrule(lr){8-10} \cmidrule(lr){11-11}
                                 & \textbf{MultiWOZ}  & \textbf{Tau-Bench} & \textbf{Avg.}      & \textbf{MultiWOZ}  & \textbf{Tau-Bench} & \textbf{Avg.}      & \textbf{MultiWOZ}   & \textbf{Tau-Bench} & \textbf{Avg.}      & \textbf{Avg.}    \\ \midrule
o1                               & \underline{4.4722} & \textbf{4.7680}    & \underline{4.6201} & 4.3623             & \underline{4.3198} & 4.3412             & \textbf{4.4179}    & \textbf{4.6091}    & \textbf{4.5135}     & \textbf{4.4916}            \\
GPT-4o                           & 3.9362             & \underline{4.6601} & 4.2982             & 4.0841             & 4.1256             & 4.1049             & 4.0326             & 4.3568             & 4.1947              & 4.1993            \\
GPT-4o-mini                      & 3.7133             & 4.4404             & 4.0769             & 3.9239             & 3.9068             & 3.9154             & 4.0285             & 4.2628             & 4.1457              & 4.0460            \\
GPT-3.5-turbo                    & 3.1984             & 4.4066             & 3.8025             & 3.5774             & 4.0506             & 3.8140             & 3.3899             & 3.5106             & 3.4503              & 3.6889            \\ 
Claude-3.5-Sonnet                & \textbf{4.6518}    & 4.6030             & \textbf{4.6274}    & \textbf{4.5447}    & 4.1315             & 4.3381             & \underline{4.3930} & \underline{4.4798} & \underline{4.4364}  & \underline{4.4673}    \\ 
Llama-3.1-405B-Instruct     & 4.3129             & 4.5663             & 4.4396             & 4.3741             & \textbf{4.4623}    & \textbf{4.4182}    & 4.1673             & 4.4469             & 4.3071              & 4.3883              \\
Llama-3.3-70B-Instruct      & 4.1749             & 3.9977             & 4.0863             & 4.1352             & 3.0003             & 3.5678             & 3.9003             & 3.8061             & 3.8532              & 3.8358             \\
Mistral-Large           & 4.2076             & 4.2553             & 4.2315             & 4.2795             & 4.3088             & 4.2942             & 4.1085             & 4.2724             & 4.1905              & 4.2387             \\
Qwen2.5-72B-Instruct             & 4.2172             & 4.4093             & 4.3133             & \underline{4.4085} & 4.4256             & \underline{4.4171} & 4.0301             & 4.2046             & 4.1174              & 4.2826              \\ \bottomrule
\end{tabular}%
  }
\caption{Turn-level Results. The best results are highlighted in \textbf{bold}, while the second-best results are \underline{underlined}.}
\label{tab: mturn-level}
\end{table*}

\begin{table*}[!t]
\centering
\resizebox{\textwidth}{!}{%
\begin{tabular}{llccccccc}
\toprule
\textbf{Ranking} & \textbf{Model} & \textbf{\convarena{}} & \textbf{Votes} & \textbf{Wins} & \textbf{Losses} & \textbf{Ties} & \textbf{Organization} & \textbf{License} \\
\midrule
1 & Claude-3.5-Sonnet & 1279.66 & 1488 & 1237 & 216 & 35 & Anthropic & Proprietary \\
2 & GPT-4o & 1107.40 & 1488 & 861 & 581 & 46 & OpenAI & Proprietary \\
3 & GPT-4o-mini & 1037.46 & 1488 & 743 & 707 & 38 & OpenAI & Proprietary \\
4 & Mistral-Large & 988.89 & 1488 & 659 & 785 & 44 & Mistral & Mistral Research \\
5 & Llama-3.1-405B-Instruct & 961.04 & 1488 & 708 & 718 & 62 & Meta & Llama \\
6 & Llama-3.3-70B-Instruct & 901.34 & 1488 & 591 & 840 & 57 & Meta & Llama \\
7 & GPT-3.5-turbo & 724.21 & 1488 & 265 & 1217 & 6 & OpenAI & Proprietary \\
\bottomrule
\end{tabular}%
}
\caption{Dialogue-level results and \convarena{} Leaderboard.}
\label{tab:elo}
\vspace{-0.1in}
\end{table*}

\subsection{Result}

Table~\ref{tab:agreements_final} compares human ratings (from the second step of human evaluation) with scores from traditional metrics, LMUnit\textsubscript{TD}, and \sysname{}, at both the turn and dialogue levels. Traditional metrics are shown only at the dialogue level, as they are computed over entire conversations. Additionally, we modified LMUnit\textsubscript{TD} to produce dialogue-level evaluations as well, details of which are provided in \Cref{app:human_eval_lmunit}. To assess alignment with human judgments, we again use Gwet’s AC1~\citep{gwet} and Randolph’s $\kappa$~\citep{randolph-kappa}. We compute the overall turn-level score by aggregating the three sub-metric scores and measuring agreement across all turns. In \Cref{tab:agreements_final}, it can be seen that \sysname{} shows stronger alignment with human judgments compared to traditional Success rate and $\tau$-Bench reward. Moreover, across both turn-level and dialogue-level evaluations, \sysname{} consistently outperforms LMUnit\textsubscript{TD}. Thus, \sysname{} offers a more reliable and human-aligned assessment of conversational quality. 

\section{Experiments} 

\subsection{Experimental Settings}
\label{sec:exp_set}

\noindent\textbf{Evaluation Instances.} To assess the performance of \sysname{}, we utilize two popular conversational dialogue benchmarks: MultiWOZ 2.4~\citep{ye2022multiwoz24} and $\tau$-Bench~\citep{yao2025taubench}, both originally evaluated using automatic metrics. We selected 100 dialogues from MultiWOZ 2.4 with human validation and used all 165 dialogues from $\tau$-Bench, covering the retail and airline domains. Finally, we report the accuracy of our LLM-based judge separately on each benchmark dataset.

\vspace{1mm}

\noindent\textbf{Models.} We evaluate 9 different LLMs including both closed-source and opened-source models: o1, GPT-4o, GPT-4o-mini, GPT-3.5-turbo (OpenAI), Llama-3.3-70B-Instruct, Llama-3.1-405B-Instruct (Meta), Claude-3.5-Sonnet (Anthropic), Mistral-Large (Mistral Research), Qwen 2.5-72B-Instruct (Alibaba). Each model is prompted to produce system responses given multi-turn dialog contexts and database records, using the same policy instructions described in \Cref{sec:turn-level}. To evaluate the generated responses, we employ GPT-4o as the default model in both the LLM-based Judge and User Simulator.

\vspace{1mm}

\noindent\textbf{User Simulators.} In our evaluation framework, we employed online user simulators to facilitate interactive assessment of dialogue agents. For experiments on MultiWOZ 2.4, we utilized the AutoTOD~\citep{xu2024autotod} user simulator and for $\tau$-Bench evaluations, we leveraged the benchmark's official user simulator as provided by the authors~\citep{yao2025taubench}.  Both simulators were powered by GPT-4o as the underlying LLM.

\subsection{Main Results}

\noindent\textbf{Turn-level Evaluation Results.} \Cref{tab: mturn-level} shows the average Likert scores (1--5) on \emph{conversation cohesion}, \emph{backend knowledge}, and \emph{policy compliance}, along with an overall average across these three dimensions. The o1 model achieves the highest overall score of 4.4916, demonstrating strong performance in policy compliance due to its advanced reasoning capabilities in following instructions. Following that, Claude-3.5-Sonnet ranks second, with Llama-3.1-405B placing third as the top open-source model. Notably, Llama-3.1-405B achieves the highest score in \emph{backend knowledge} (4.4182), outperforming proprietary models. \Cref{tab: mturn-level} indicates that while certain models perform well in cohesion and factual correctness, they may fail to strictly enforce policies. This highlights the need for a comprehensive evaluation in multiple dimensions.

\begin{figure*}[!t]
\includegraphics[width=\linewidth]{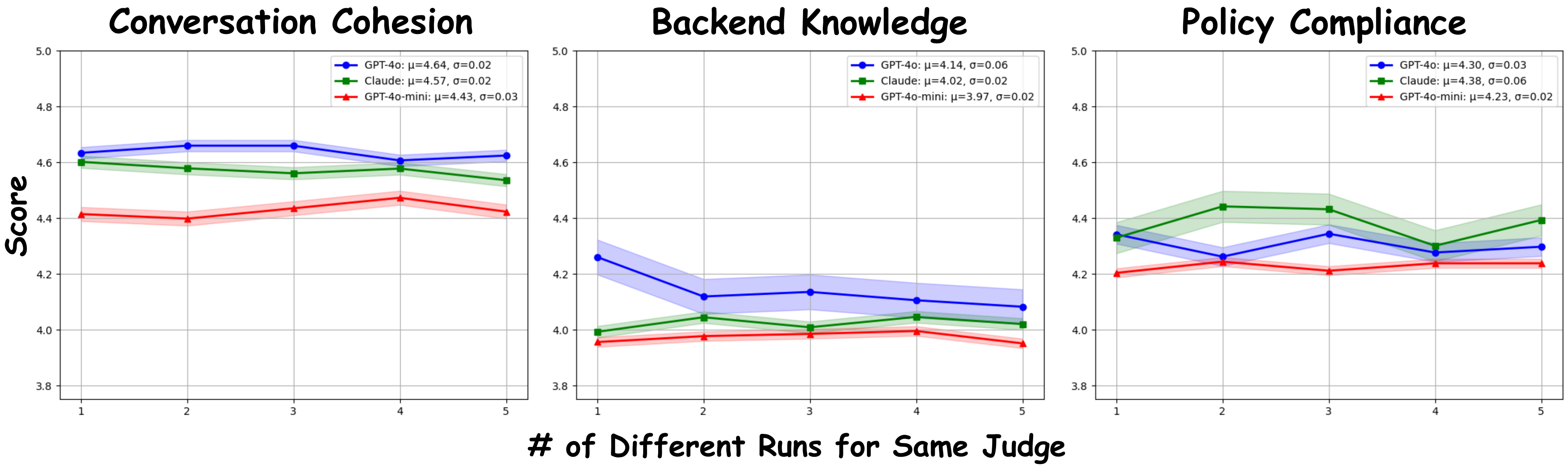}
\caption{\textbf{LLM Judge Consistency Across Multiple Runs.} Evaluation results for three metrics using the same LLM judge over five runs: conversation cohesion (left), backend knowledge (middle), and policy compliance (right).}
\vspace{-3mm}
\label{fig:ablation1}
\end{figure*}

\noindent\textbf{Dialogue-level Evaluation Results.} \Cref{tab:elo} summarizes the pairwise Elo rankings under our \convarena{}. Claude-3.5-Sonnet dominates the rankings with an Elo rating of 1514.42, winning 571 out of 600 head-to-head matchups in dialogue-level evaluations. This aligns with its strong turn-level performance, indicating its effectiveness in end-to-end conversations. Although Mistral-Large did not rank at the very top in turn-level scores, it unexpectedly places second in the dialogue-level evaluation (Elo = 1068.67), implying that it manages to recover from localized errors and provides more user-friendly end-to-end interactions. GPT-4o and Meta-Llama-3.1-405B Instruct both maintain solid Elo ratings (1055.40 and 1020.36, respectively), consistent with their strong turn-level performance. In contrast, GPT-3.5-turbo 
sits at the bottom, aligning with its lower per-turn scores.

\subsection{Ablation Studies}

\noindent\textbf{\sysname{} demonstrates consistent scoring in multiple runs.} A common concern with LLM-based evaluation is the potential variability in assessment outcomes. To investigate this, we conducted a reliability analysis by \textit{running the same judge five times} (GPT-4o) on identical responses from three agents (Claude-3.5-Sonnet, GPT-4o-mini, and GPT-4o) on $\tau$-Bench. \Cref{fig:ablation1} visualizes score variation over multiple runs together with standard deviations. The nearly flat curves and low standard deviations demonstrate that \sysname{} yields highly consistent judgments, underscoring the robustness of our evaluation protocol.

\vspace{1mm}

\noindent\textbf{\sysname{} scores are robust to different LLM judges.} 
The \sysname{} scoring remains stable across diverse LLM judges. In \Cref{fig:ablation2} (\Cref{app:abl2}), we compare different agents with three different LLM judges. We observe that all three judges produce closely aligned absolute scores. 
Refer to \Cref{app:abl2} for further details.




\section{Related Work}

\subsection{LLM-Based TOD Evaluation}
Recent research on TOD has leveraged LLMs through various ways, including fine-tuning~\citep{su-etal-2022-multi} and prompting strategies~\citep{hu2022context}, with particular emphasis on challenges related to database access and logical reasoning~\citep{hudecek2023arellmstod}. With these challenges, comprehensive evaluation remains a key bottleneck in TOD research. While early works—such as \citet{PARADISE}—established the foundational evaluation paradigms, the advancement of TOD evaluation metrics has lagged behind that of open-domain dialogue systems, which has introduced diverse approaches, such as reference-based~\citep{ADEM} and reference-free~\citep{FlowScore} methods. In the TOD domain, automated evaluation metrics often fail to capture nuanced multi-turn errors~\citep{mehri-etal-2019-structured}. To address these limitations, \sysname{} introduces multi-perspective LLM-based evaluation metrics specifically tailored to TOD scenarios.

\subsection{Multi-Turn TOD Evaluation}
The inherent multi-turn nature of TOD presents additional challenges for evaluation, largely due to the error accumulation across turns~\citep{10.1145/3240508.3240605, liao-etal-2021-dialogue}. Although prior studies have explored both turn~\citep{engelbrech2009userhmm, ryuichiro2010issues, ultes2014interaction, georgila2024comparing} and dialogue level evaluation approaches~\citep{bodigutla-etal-2020-joint, xu2024autotod}, developing a comprehensive evaluation metric capable of assessing the overall interaction quality remains an open problem, particularly for TOD systems built on LLMs. To resolve this gap, \sysname{} provides a systematic framework designed to evaluate TOD interactions at both the turn and dialogue levels, providing a more holistic assessment for TOD systems based on LLMs. 
\section{Conclusion}
We present \sysname{}, a simple yet powerful framework for TOD evaluation that combines fine-grained turn-level checks with a holistic dialogue-level ranking. By adopting an LLM-as-judge paradigm, \sysname{} goes beyond standard metrics to reveal subtle yet critical errors, such as inconsistent database usage and policy violations, which often remain undetected by final-turn or dialogue-level summaries. Through Elo-based ranking and targeted turn-level scoring, our experiments on MultiWOZ~2.4 and \(\tau\)-Bench demonstrate \sysname{}’s alignment with human judgments. This work opens a new path for LLM-driven TOD evaluation—one that is both flexible and transparent—ensuring greater accountability and accuracy in developing next-generation dialogue systems. We intend to release our framework, system responses, and human evaluations to foster reproducibility and community adoption.
\section{Limitations} 

While metrics in \sysname{} cover core aspects occurring in general TOD scenarios, it still remains open questions to design more flexible, fine-grained evaluation metrics that can cover diverse scenarios during multi-turn interactions. Furthermore, practitioners should consider that the performance of LLM-based evaluation can be improved when appending qualified few-shot demonstrations or tailored scoring rubrics in specific service domains. Lastly, it should be noted that conventional evaluation metrics are still useful to evaluate TOD in specific aspects, thus \sysname{} should be used alongside existing metrics in a complementary relationship.

\section{Ethics Statement} 
We conduct our experiments using the publicly available MultiWOZ and $\tau$-Bench datasets, adhering fully to their terms of use. Since we employ LLMs to generate evaluations with justifications, the risk of producing harmful, biased, or discriminatory statements is minimal. However, we acknowledge the potential ethical concerns associated with this work.
\section*{Acknowledgement}
This work was supported in part by Other Transaction award HR0011249XXX from the U.S. Defense Advanced Research Projects Agency (DARPA) Friction for Accountability in Conversational Transactions (FACT) program and has benefited from the Microsoft Accelerate Foundation Models Research (AFMR) grant program, through which leading foundation models hosted by Microsoft Azure and access to Azure credits were provided to conduct the research.

\bibliography{custom}

\clearpage
\newpage
\appendix
\label{sec:appendix}

\section{Additional Human Evaluation Details}
\label{app:human_eval}
This section provides the additional details related to human evaluation described in Section~\ref{sec:human_eval}.

\subsection{Human Evaluation Setup}
\label{app:human_eval_setup}
\begin{figure*}[!t]
\includegraphics[width=1.0\linewidth]{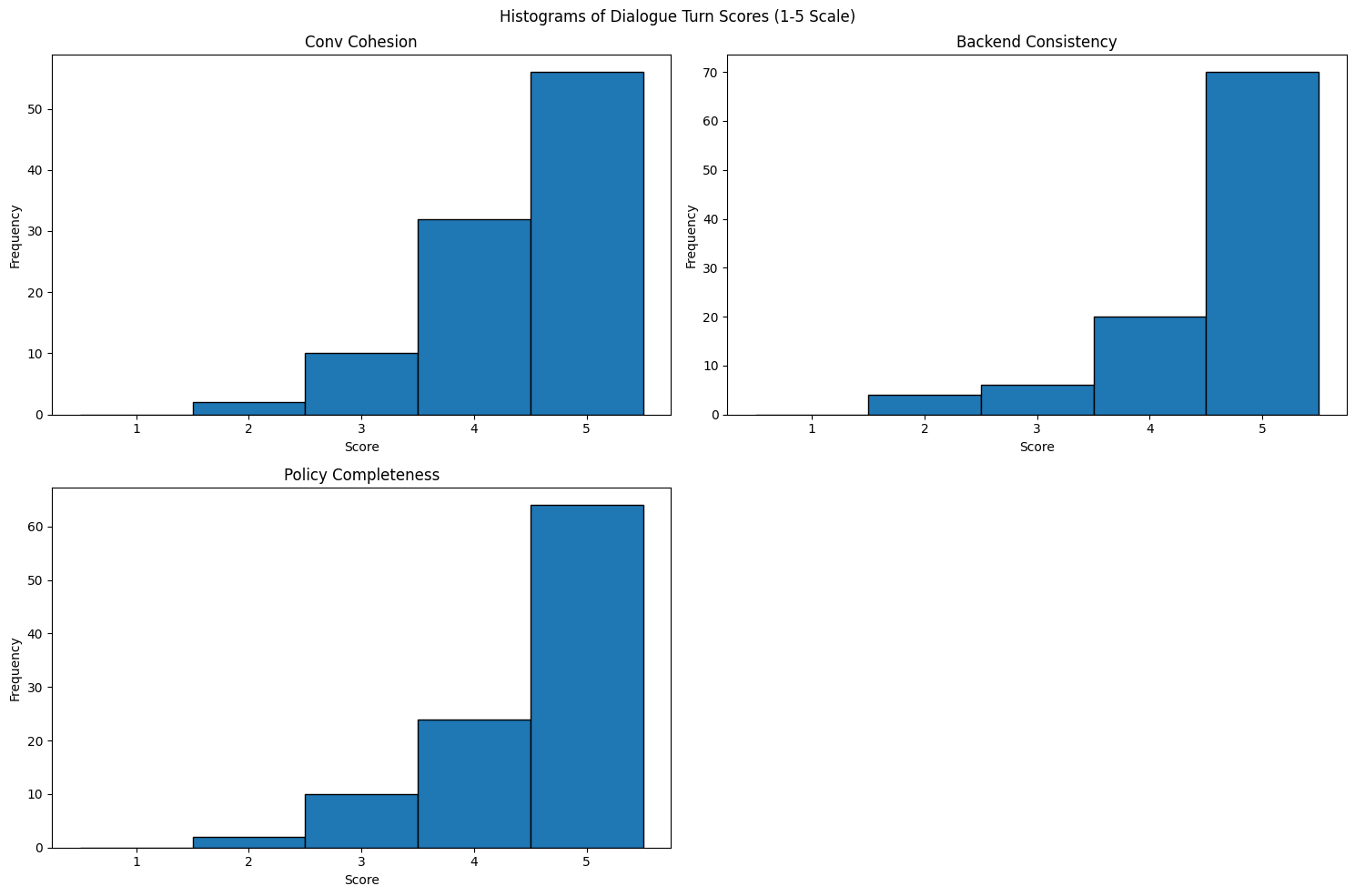}
\caption{\textbf{Human Evaluation Scores Distribution} This is the distribution of all scores from 10 annotators during the inter-annotator agreement evaluation stage, separated by the three metrics. For all three TD-Eval metrics, the vast majority of LLM responses were classified as 4 (Good) or 5 (Very Good), resulting in class distribution imbalance.}
\vspace{-3mm}
\label{fig:score-distribution}
\end{figure*}

\begin{figure*}[!t]
\centering
\includegraphics[width=0.8\linewidth]{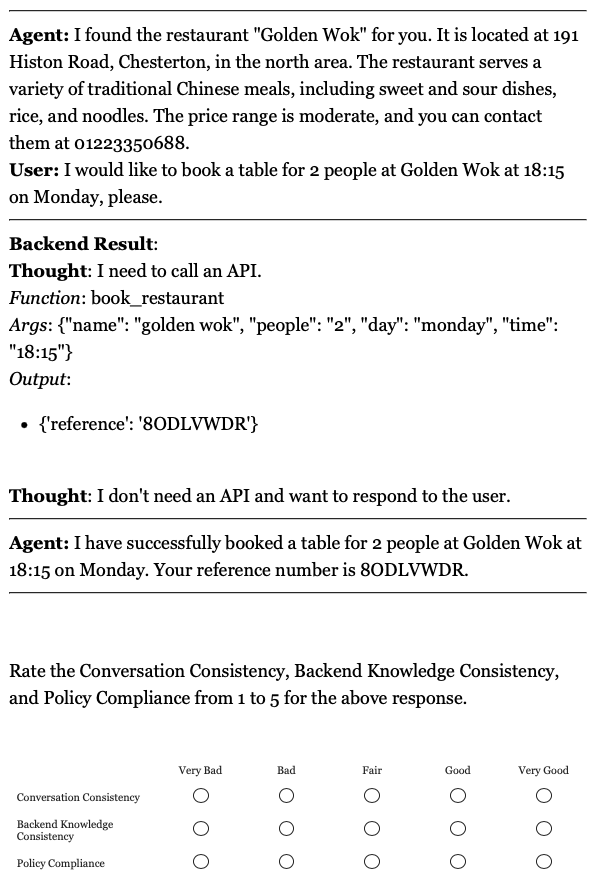}
\caption{\textbf{Human Evaluation Interface:} This is the interface showing an example evaluation question in the human evaluation study. We conduct the evaluation on Qualtrics.}
\label{fig:qualtrics-eval}
\end{figure*}

Human evaluation was conducted through Qualtrics, an online survey platform. Ten annotators were recruited for evaluation via convenience sampling, where the authors selected participants from their personal and professional networks. We include a snapshot of the survey interface that was used for human evaluation in Figure \ref{fig:qualtrics-eval}. We provide detailed instructions to human annotators, shown in Figure~\ref{fig:eval-instructions}, as reference when evaluating conversations. We observed a strong skew in the collected ratings, with nearly 90\% of scores falling in the 4–5 range, which is shown in Figure \ref{fig:score-distribution}. 

\subsection{LMUnit Scoring}
\label{app:human_eval_lmunit}
The natural language unit tests and API calls~\footnote{\href{https://contextual.ai/lmunit/}{contextual.ai/lmunit/}} to compute the LMUnit scores is shown in Figure~\ref{fig:lmunit-prompts}. LMUnit~\cite{lmunit} scores are returned as floating point numbers with decimal points, while Traditional and \sysname{} scoring is integer-based. To address this, LMUnit scores are rounded to the nearest integer after averaging the unit test scores. Turn-level scores are calculated based on the three sub-metrics to calculate agreement with \sysname{} scores. In the dialogue level, there is only one score that broadly measures goal completion performance of the TOD agent for the overall conversation. To mirror traditional metrics, we change LMUnit to evaluate based on goal completion rather than the original sub-metrics. This score is also calculated on a [1, 5] scale. 

\subsection{Additional Results}
In this section, we perform an extended analysis of the turn and dialogue-level results in Table~\ref{tab:agreements_final} to find more fine-grained human agreement results of each evaluation metric. Table \ref{tab:agreements_abl_turn} shows the turn-level agreement results split by \sysname{} sub-metrics. Table \ref{tab:agreements_abl_dial} lays out the dialogue-level agreement results split by the dataset (MultiWOZ or $\tau$-bench). Besides the traditional Success rate, we also report AutoTOD's online Success rate~\cite{xu2024autotod} that operates on raw, non-delexicalized responses. In Table \ref{tab:agreements_abl_dial}, Success\textsubscript{OR} refers to the traditional Success rate computed with delexicalized responses, while Success\textsubscript{AT} denotes the online Success rate. For both turn and dialogue level, \sysname{} consistently shows greater alignment with human judgment.

\begin{table}[h!]
\centering
\begin{small}

\begin{tabular}{l m{1.6cm} c c} 
\toprule
\textbf{Metric} & \textbf{Sub-Metric}  & \textbf{Gwet AC1}  & \textbf{Randolph's $\kappa$} \\ \midrule
\multirow{6}{*}{LMUnit\textsubscript{TD}}
& Conversation Cohesion & 0.45 & 0.43  \\
& Backend Knowledge & 0.42 & 0.40  \\
& Policy Compliance & 0.41 & 0.39  \\
& Overall & 0.43 & 0.41 \\ 
\midrule
\multirow{6}{*}{\sysname{}}
& Conversation Cohesion & 0.56 & 0.51  \\
& Backend Knowledge & 0.55 & 0.50  \\
& Policy Compliance & 0.56 & 0.50  \\
& Overall & 0.56 & 0.50 \\ 
\bottomrule
\end{tabular}%
\caption{Extended comparison of turn-level performance of LMUnit\textsubscript{TD} and \sysname{} with human ratings, split by \sysname{} sub-metrics.}
\vspace{-0.1in}
\label{tab:agreements_abl_turn}
\end{small}
\end{table}

\begin{table}[h!]
\centering
\begin{small}

\begin{tabular}{l l c c} 
\toprule
\textbf{Dataset} & \textbf{Metric}  & \textbf{Gwet AC1}  & \textbf{Randolph's $\kappa$} \\ \midrule
\multirow{4}{*}{MultiWOZ}
& Success\textsubscript{OR} & 0.40 & 0.33  \\
& Success\textsubscript{AT} & 0.47 & 0.34  \\
& LMUnit\textsubscript{TD} & 0.49 & 0.47  \\
& TD-EVAL & 0.63 & 0.59 \\ 
\midrule
\multirow{3}{*}{$\tau$-Bench}
& Reward & 0.43 & 0.33 \\
& LMUnit\textsubscript{TD} & 0.34 & 0.30 \\
& TD-EVAL & 0.47 & 0.42 \\
\bottomrule
\end{tabular}%
\caption{Extended comparison of dialogue-level. Scores are split up by dataset type. Success\textsubscript{OR} indicates the traditional Success rate computed with delexicalized responses, while Success\textsubscript{AT} denotes the online Success rate without any delexicalization of system response.}
\label{tab:agreements_abl_dial}
\end{small}
\end{table}

\section{Additional Ablation Study}
\label{app:abl2}
\begin{figure*}[!t]
\includegraphics[width=1.0\linewidth]{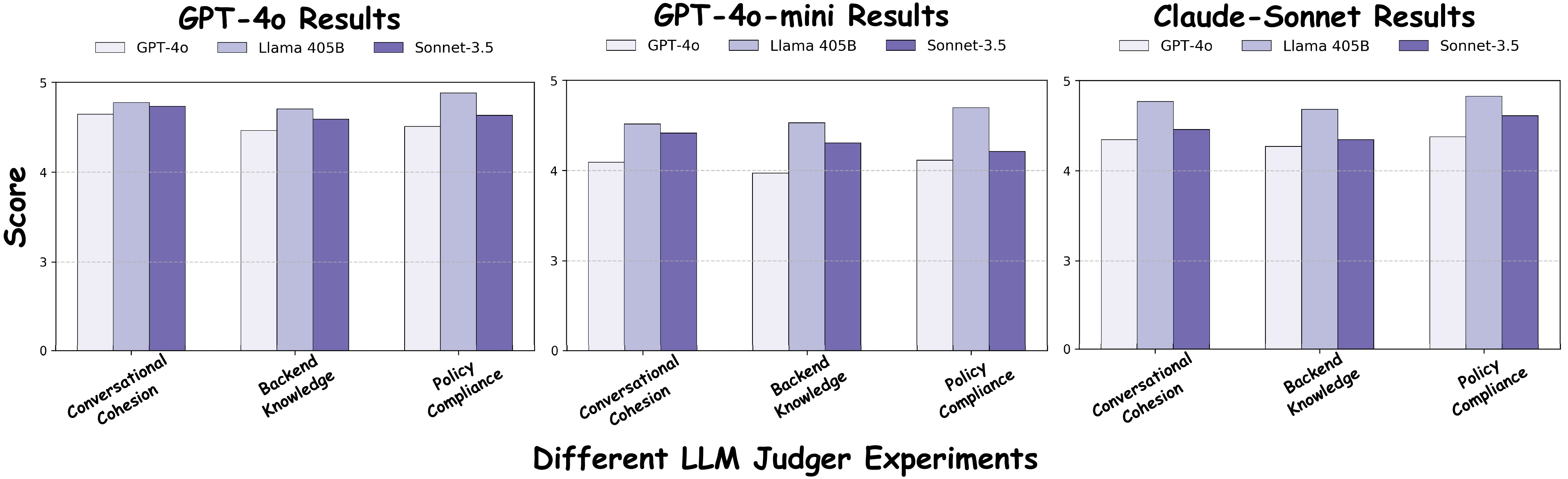}
\caption{\textbf{Different LLM Judge Experiments.} Evaluation scores of GPT-4o (left), GPT-4o-mini (center), and Claude-Sonnet-3.5 (right) under different LLM judges (GPT-4o, Llama-405B, and Claude-3.5-Sonnet).}
\label{fig:ablation2}
\end{figure*}

For this ablation study, we compare GPT-4o, GPT-4o-mini, and Claude-3.5-Sonnet with three different LLM judges (GPT-4o, Llama-405B, and Claude-3.5-Sonnet), shown in Figure \ref{fig:ablation2}. 
We observe that all three judges produce closely aligned absolute scores. Furthermore, the relative ranking of the models evaluated remains consistent between different judges. Interestingly, Llama-405B sometimes assigns higher scores than its proprietary counterparts, possibly reflecting different reasoning preferences. 
Overall, our results demonstrate that the scoring remains stable across diverse LLM judges.

\begin{figure*}[ht]
    \begin{tcolorbox}
    \begin{tabular}{p{\linewidth}}
    \small
    Thank you for participating in this annotation task. We have provided 8 dialogues where an AI agent is responding to user requests, and we want you to evaluate the responses. This process should take roughly 45-60 minutes. Progress will be saved automatically, so you can complete it in multiple sessions, but \textbf{make sure you are using the same browser each time}. Please rate individual dialogue responses of AI agents from 1 (worst) to 5 (best) on the following qualities: \textbf{Conversation Consistency}, \textbf{Backend Knowledge Consistency}, and \textbf{Policy Compliance}, the metric definitions are below. In addition, please rate the full dialogue in terms of task completion and response coherence. \\\small 
    \\\small
    \textbf{Conversation Consistency} \\\small
    How much an agent's response align with the context of conversation context. 
    \begin{itemize}[noitemsep,topsep=0pt,parsep=0pt,partopsep=0pt]
        \item \textbf{Relevance}: The response directly relates to the dialogue history and the current user query.
        \item \textbf{Topic Consistency}: The response remains on-topic with the dialogue history and the user query.
        \item \textbf{Coherence}: The response logically continues the progression of the dialogue.
    \end{itemize}
    \\\small
    \textbf{Scoring Scale} 
    \begin{enumerate}[noitemsep,topsep=0pt,parsep=0pt,partopsep=0pt]
        \item \textbf{Very Good}: Response is completely consistent with the previous conversation context, with no inconsistencies or errors.
        \item \textbf{Good}: Response is mostly consistent with the context. Only minor improvements are needed.
        \item \textbf{Fair}: Response is somewhat consistent but contains noticeable inconsistencies or lacks depth in addressing the context.
        \item \textbf{Bad}: Response shows limited consistency with the conversation context and requires significant improvement.
        \item \textbf{Very Bad}: Response is incoherent or completely inconsistent with the conversation context.
    \end{enumerate} 
    \\\small
    \textbf{Backend Knowledge Consistency} \\\small
    How well an agent's response aligns with information provided by backend database results. 
    \begin{itemize}[noitemsep,topsep=0pt,parsep=0pt,partopsep=0pt] 
        \item \textbf{Accuracy}: The response directly reflects the information in the database results.
        \item \textbf{Topic Consistency}: The response stays on-topic with the database results and the dialogue context.
        \item \textbf{Coherence}: The response logically incorporates and progresses based on the database results.
    \end{itemize} 
    \\\small
    \textbf{Scoring Scale} 
    \begin{enumerate}[noitemsep,topsep=0pt,parsep=0pt,partopsep=0pt]
        \item \textbf{Very Good}: Response is completely consistent with the database results, with no inconsistencies or errors.
        \item \textbf{Good}: Response is mostly consistent with the database results. Only minor improvements are needed.
        \item \textbf{Fair}: Response is sufficiently consistent with the database results but contains noticeable inconsistencies or lacks depth in addressing the results.
        \item \textbf{Bad}: Response shows limited consistency with the database results and requires significant improvement.
        \item \textbf{Very Bad}: Response is incoherent or completely inconsistent with the database results.
    \end{enumerate}
    \\\small
    \textbf{Policy Compliance} \\\small
    How well an agent's response adheres to the expected policy protocol. 
    \begin{itemize}[noitemsep,topsep=0pt,parsep=0pt,partopsep=0pt]
        \item \textbf{Number of Suggestions}: Providing suggestions only when the database results are small enough to do so.
        \item \textbf{Information Gathering}: Requesting required, relevant information (slots) from the user before offering suggestions or booking services.
        \item \textbf{Appropriate Timing}: Avoiding premature actions, such as making a booking or suggesting a service too early in the conversation.
        \item \textbf{Alignment with Policy}: Avoiding actions that do not align with the suggested flow of interaction in the policy, when available.
    \end{itemize}
    \\\small
    \textbf{Expected Policy} \\\small
    The chatbot response should depend on the database results and dialogue history: 
    \begin{itemize}[noitemsep,topsep=0pt,parsep=0pt,partopsep=0pt]
        \item If the database results return a number larger than 10: Indicate the number of entries that match the user's query and request additional information if needed to narrow down the results.
        \item If the database results return values less than 10: If vital details are missing, request additional information. Otherwise, provide the relevant entries to the user
    \end{itemize}
    \\\small
    \textbf{Scoring Scale} 
    \begin{enumerate}[noitemsep,topsep=0pt,parsep=0pt,partopsep=0pt]
        \item \textbf{Very Good}: Response fully follows policy protocol with no errors or omissions.
        \item \textbf{Good}: Response mostly follows policy protocol, with only minor room for improvement.
        \item \textbf{Fair}: Response sufficiently follows policy protocol but has clear areas where it could improve in completeness or timing.
        \item \textbf{Bad}: Response does not adequately follow policy protocol, though there may be partial adherence.
        \item \textbf{Very Bad}: Response fails to follow policy protocol and is incomplete or incoherent.
    \end{enumerate}
    \end{tabular}
    \end{tcolorbox}
    \caption{The evaluation instructions provided as reference for human evaluation on Qualtrics platform.}
    \label{fig:eval-instructions}
\end{figure*}
\begin{figure*}[ht]
    \begin{tcolorbox}
    \begin{tabular}{p{\linewidth}}
    \textbf{Turn-Level Unit Tests for Conversation Cohesion}
    
    \begin{enumerate}[itemsep=0pt]
        \item \textbf{Accuracy}: Does the response directly relate to the dialogue history and the current user query? 
        \item \textbf{Topic Consistency}: Does the response remain on-topic with the dialogue history and the user query? 
        \item \textbf{Coherence}: Does the response logically continue the progression of the dialogue?
    \end{enumerate}

    \textbf{Turn-Level Unit Tests for Backend Knowledge Consistency}
    \begin{enumerate}[itemsep=0pt]
        \item \textbf{Accuracy}: Does the response accurately reflect the information in the database results? 
        \item \textbf{Topic Consistency}: Does the response stay on-topic with the database results and the dialogue context?
        \item \textbf{Coherence}: Does response logically incorporate and progress based on the database results?
    \end{enumerate}

    \textbf{Turn-Level Unit  Tests for Policy Compliance}
    \begin{enumerate}[itemsep=0pt]
        \item \textbf{Number of Suggestions}: Does the response provide suggestions only when the database results are few enough to do so? 
        \item \textbf{Information Gathering}: Does the response request required, relevant information from the user before offering suggestions or booking services?
        \item \textbf{Appropriate Timing}: Does the response avoid premature actions (i.e. make a booking or suggest a service) too early in the conversation, before the necessary information is gathered?
    \end{enumerate}

    \textbf{API call template for turn-level score} \\
    \{ \\
        \quad ``query": [CONV\_HISTORY] \textbackslash n ``Database Result:" [DB\_RESULT] \\
        \quad ``response": [AGENT\_RESPONSE] \\
        \quad ``unit\_test": [UNIT\_TEST\_QUESTION] \\
    \} \\ \\

    \textbf{Dialogue-Level Unit Test for Goal Completion}
    \begin{enumerate}[itemsep=0pt]
        \item \textbf{Goal Completion:} Does the conversation achieve it's intended goal?
    \end{enumerate}

    \textbf{API call template for dialogue-level score} \\
    \{ \\
        \quad ``query": ``Goal: " [CONV\_GOAL]  \\
        \quad ``response": [CONV\_HISTORY] \\
        \quad ``unit\_test": ``Does the conversation achieve it's intended goal?" \\
    \} \\
    
    
    \end{tabular}
        
\end{tcolorbox}
    \caption{Natural language unit tests for computing the LMUnit\textsubscript{TD} scores at the turn and dialogue level. For the turn level, we calculate scores for the three sub-metrics. Scores from each unit test are between [1, 5]. For the dialogue metric, we only ask whether the conversation achieved it's intended goal, which is supplied in both MultiWOZ and $\tau$-bench datasets. We provide the API call template to LMUnit as additional context.}
    \label{fig:lmunit-prompts}
\end{figure*}
\begin{figure*}[ht]
    \begin{tcolorbox}
    \begin{tabular}{p{\linewidth}}
    Evaluate the \textbf{conversation consistency} of the following task-oriented dialogue chatbot response on a 5-point scale from "Very Bad" to "Very Good". \\
    The prompt will include the \textbf{dialogue history}, \textbf{current user query}, \textbf{database results}, \textbf{chatbot response}.  \\
     \\
    \textbf{Conversation Consistency Definition} \\
    Conversation consistency refers to the degree to which the chatbot's response aligns with the context of the conversation, including: 
    \begin{itemize}[itemsep=0pt]
        \item \textbf{Relevance}: The response directly relates to the dialogue history and the current user query.
        \item \textbf{Topic Consistency}: The response remains on-topic with the dialogue history and the user query.
        \item \textbf{Coherence}: The response logically continues the progression of the dialogue. 
    \end{itemize}
    \textbf{Scoring Guide}: 
    \begin{itemize}[itemsep=0pt]
        \item \textbf{Very Good (5)}: Response is completely consistent with the previous conversation context, with no inconsistencies or errors.
        \item \textbf{Good (4)}: Response is mostly consistent with the context. Only minor improvements are needed. 
        \item \textbf{Fair (3)}: Response is somewhat consistent but contains noticeable inconsistencies or lacks depth in addressing the context. 
        \item \textbf{Bad (2)}: Response shows limited consistency with the conversation context and requires significant improvement. 
        \item \textbf{Very Bad (1)}: Response is incoherent or completely inconsistent with the conversation context. 
    \end{itemize}
    \textbf{Always include the score first, then a rationale on a new line. The rationale should be at most 2 sentences. Follow the template below}: \\
    Score: [YOUR SCORE NUMBER HERE] \\
    Justification: [YOUR RATIONALE HERE] \\
     \\
    \textbf{Evaluate the conversation consistency of the following dialogue response} \\
    \textit{Dialogue History} \\
    \{dialogue\_history\} \\
    \textit{User Query} \\
    \{user\_query\} \\
    \textit{Database Results} \\
    \{db\_result\} \\
    \textit{Chatbot Response} \\
    \{agent\_response\} \\
    \end{tabular}
        
\end{tcolorbox}
    \caption{The GPT-4o prompt used for evaluating Conversational Consistency on MultiWOZ 2.4.}
    \label{fig:conversational-consistency}
\end{figure*}

\begin{figure*}[ht]
    \begin{tcolorbox}
    \begin{tabular}{p{\linewidth}}
    Evaluate the \textbf{backend knowledge consistency} of the following task-oriented dialogue chatbot response on a 5-point scale from "Very Bad" to "Very Good". \\
    The prompt will include the \textbf{dialogue history}, \textbf{current user query}, \textbf{database results}, \textbf{chatbot response}.  \\
     \\
    \textbf{Backend Knowledge Consistency Definition} \\
    Backend knowledge consistency refers to how well the chatbot's response aligns with the information provided in the policy or database results, considering:
    \begin{itemize}[itemsep=0pt]
        \item \textbf{Accuracy}: The response accurately reflects the information in the database results. 
        \item \textbf{Topic Consistency}: The response stays on-topic with the database results and the dialogue context. 
        \item \textbf{Coherence}: The response logically incorporates and progresses based on the database results.
    \end{itemize}
    \textbf{Scoring Guide}:
    \begin{itemize}[itemsep=0pt]
        \item \textbf{Very Good (5)}: Response is completely consistent with the database results, with no inconsistencies or errors. 
        \item \textbf{Good (4)}: Response is mostly consistent with the database results. Only minor improvements are needed. 
        \item \textbf{Fair (3)}: Response is sufficiently consistent with the database results but contains noticeable inconsistencies or lacks depth in addressing the results. 
        \item \textbf{Bad (2)}: Response shows limited consistency with the database results and requires significant improvement. 
        \item \textbf{Very Bad (1)}: Response is incoherent or completely inconsistent with the database results. 
    \end{itemize}

    \textbf{Always include the score first, then a rationale on a new line. The rationale should be at most 2 sentences. Follow the template below}: \\
    Score: [YOUR SCORE NUMBER HERE] \\
    Justification: [YOUR RATIONALE HERE] \\
     \\
    \textbf{Evaluate the backend knowledge consistency of the following dialogue response} \\
    \textit{Dialogue History} \\
    \{dialogue\_history\} \\
    \textit{User Query} \\
    \{user\_query\} \\
    \textit{Database Results}  \\
    \{db\_result\}  \\
    \textit{Chatbot Response}  \\
    \{agent\_response\}  \\
    \end{tabular}
        
\end{tcolorbox}
    \caption{The GPT-4o prompt used for evaluating Backend Knowledge on MultiWOZ 2.4.}
    \label{fig:backend-knowledge}
\end{figure*}
\begin{figure*}[ht]
    \begin{tcolorbox}
    \begin{tabular}{p{\linewidth}}
    \small
    Evaluate the \textbf{policy compliance} of the following task-oriented dialogue chatbot response on a 5-point scale from "Very Bad" to "Very Good". The prompt will include the \textbf{policy protocol}, \textbf{dialogue history}, \textbf{current user query}, \textbf{database results}, \textbf{chatbot response}.  \\\small
    \\[-8pt] \small
    \textbf{Policy Compliance Definition} \\\small
    Policy compliance refers to how well the chatbot adheres to the expected policy protocol, specifically: 
    \begin{itemize}[noitemsep,topsep=0pt,parsep=0pt,partopsep=0pt]
        \item \textbf{Number of Suggestions}: Providing suggestions only when the database results are few enough to do so. 
        \item \textbf{Information Gathering}: Requesting required, relevant information (slots) from the user before offering suggestions or booking services. 
        \item \textbf{Appropriate Timing}: Avoiding premature actions, such as making a booking or suggesting a service too early in the conversation, before the necessary information is gathered from the user. 
        \item \textbf{Alignment with Policy}: Avoiding actions that do not align with the suggested flow of interaction in the policy, when available.
    \end{itemize}
    \\[-8pt] \small
    \textbf{Domain Possible Slots (May not be exhaustive)} \\\small
    The current predicted domain for this turn in the conversation is "Restaurant". You should check if domain-relevant slots have been filled in the current conversation. The possible slots for all domains are shown below (these may not be totally exhaustive): \\\small
    \\[-8pt] \small
    \textit{Restaurant} 
    \begin{itemize}[noitemsep,topsep=0pt,parsep=0pt,partopsep=0pt]
        \item bookpeople: the number of people included in the restaurant reservation 
        \item booktime: the time for the reservation at the restaurant 
        \item food: the type of cuisine the restaurant serves 
        \item name: the name of the restaurant 
        \item pricerange: the price range of the restaurant 
    \end{itemize}

    \\[-8pt] \small
    \textbf{Scoring Guide}: 
    \begin{itemize}[noitemsep,topsep=0pt,parsep=0pt,partopsep=0pt]
        \item \textbf{5 (Very Good)}: Response fully follows policy protocol with no errors or omissions. 
        \item \textbf{4 (Good)}: Response mostly follows policy protocol, with only minor room for improvement. 
        \item \textbf{3 (Fair)}: Response sufficiently follows policy protocol but has clear areas where it could improve in completeness or timing. 
        \item \textbf{2 (Bad)}: Response does not adequately follow policy protocol, though there may be partial adherence. 
        \item \textbf{1 (Very Bad)}: Response fails to follow policy protocol and is incomplete or incoherent. 
    \end{itemize}

    \\[-8pt] \small
    \textbf{Always include the score first, then a rationale on a new line. The rationale should be at most 2 sentences. Follow the template below}: \\\small
    Score: [YOUR SCORE NUMBER HERE] \\\small
    Justification: [YOUR RATIONALE HERE] \\\small
    \\[-8pt] \small
    \textbf{Policy Protocol} \\\small
    The chatbot response should depend on the database results and dialogue history: 
    \begin{enumerate}[noitemsep,topsep=0pt,parsep=0pt,partopsep=0pt]
        \item If the database results return a number: Indicate the number of entries that match the user's query and request additional information if needed to narrow down the results. 
        \item If the database results return values: If vital details are missing based on the dialogue history, request additional information. Otherwise, provide the relevant entries to the user. 
    \end{enumerate}

    \\[-8pt] \small
    \textbf{Evaluate the policy completeness of the following dialogue response} \\\small
    \textit{Dialogue History} \\\small
    \{dialogue\_history\} \\\small
    \textit{User Query} \\\small
    \{user\_query\} \\\small
    \textit{Database Results} \\\small
    \{db\_result\} \\\small
    \textit{Chatbot Response} \\\small
    \{agent\_response\}
    \end{tabular}
        
\end{tcolorbox}
    \caption{The GPT-4o prompt used for evaluating Policy Compliance Consistency on MultiWOZ 2.4 restaurant domain.}
    \label{fig:policy-compliance}
\end{figure*}
\begin{figure*}[ht]
    \begin{tcolorbox}
    \begin{tabular}{p{\linewidth}}
    Compare these two AI assistant conversations and determine which one is better. \\
    Consider the following aspects:
     \begin{enumerate}
        \item \textbf{Conversation Consistency}:
        \begin{itemize}
            \item \textbf{Relevance}: The response directly relates to the dialogue history and the current user query. 
            \item \textbf{Topic Consistency}: The response remains on-topic with the dialogue history and the user query. 
            \item \textbf{Coherence}: The response logically continues the progression of the dialogue. 
        \end{itemize}
        \item \textbf{Backend Knowledge Consistency}:
        \begin{itemize}
            \item \textbf{Accuracy}: The response accurately reflects the information in the database results. 
            \item \textbf{Topic Consistency}: The response stays on-topic with the database results and the dialogue context. 
            \item \textbf{Coherence}: The response logically incorporates and progresses based on the database results. 
        \end{itemize}
        \item \textbf{Policy Compliance}: 
        \begin{itemize}
            \item \textbf{Number of Suggestions}: Providing suggestions only when the database results are few enough to do so. 
            \item \textbf{Information Gathering}: Requesting required, relevant information (slots) from the user before offering suggestions or booking services. 
            \item \textbf{Appropriate Timing}: Avoiding premature actions, such as making a booking or suggesting a service too early in the conversation. 
            \item \textbf{Policy Protocol}: The chatbot response should depend on the database results and dialogue history: 
            \begin{enumerate}
                \item If the database results return a number: Indicate the number of entries that match the user's query and request additional information if needed to narrow down the results. 
                \item If the database results return values: If vital details are missing based on the dialogue history, request additional information. Otherwise, provide the relevant entries to the user. 
            \end{enumerate}
        \end{itemize}
     \end{enumerate}
    Conversation A: \\
    \{conv\_a\_formatted\} \\
     \\
    Conversation B: \\
    \{conv\_b\_formatted\} \\
     \\
    Which conversation was better? Answer with only: \\
    CONVERSATION\_A if Conversation A was better \\
    CONVERSATION\_B if Conversation B was better \\
    EQUAL if they were roughly equivalent \\
    \end{tabular}
        
\end{tcolorbox}
    \caption{The GPT-4o prompt used for dialogue-level evaluation in \convarena{}.}
    \label{fig:elo-prompt}
\end{figure*}

\vspace{1mm}

\end{document}